\documentclass{article}

\usepackage[final]{nips_OptOpt}

\usepackage[utf8]{inputenc} 
\usepackage[T1]{fontenc}    
\usepackage{hyperref}       
\usepackage{url}            
\usepackage{booktabs}       
\usepackage{amsmath,amssymb}
\usepackage{amsfonts}       
\usepackage{nicefrac}       
\usepackage{microtype}      

\usepackage[usenames,dvipsnames]{xcolor}
\usepackage{graphicx}
\usepackage{tikz,pgfplots}
\pgfplotsset{compat=newest}

\usepackage{algorithm}
\usepackage{algorithmic}

\usepackage{xspace}
\DeclareRobustCommand{\eg}{e.g.,\@\xspace}
\DeclareRobustCommand{\ie}{i.e.,\@\xspace}
\DeclareRobustCommand{\wrt}{w.r.t.\@\xspace}
\DeclareRobustCommand{\wp}{w.p.\@\xspace}

\usepackage{url} 

\usepackage{amsthm}
\newtheorem{theorem}{Theorem}
\newtheorem{lemma}[theorem]{Lemma}

\theoremstyle{remark}

\theoremstyle{definition}

\newcommand{\vtheta}{\boldsymbol{\theta}}
\newcommand{\transpose}[1]{{#1}^{\texttt{T}}}
\newcommand{\lstep}{\eta}
\newcommand{\vlstep}{\boldsymbol{\eta}}
\newcommand{\lstepf}[2][]{\lstep#1\left({#2}\right)}
\newcommand{\vlstepf}[2][]{\vlstep#1\left({#2}\right)}
\newcommand{\realspace}{\mathbb{R}}
\newcommand{\imp}{\Delta\loss}
\newcommand{\eimp}{\Delta\loss^n}
\newcommand{\simpimp}{\widehat{\Delta}\loss^n}

\DeclareMathOperator*{\argmax}{arg\,max}

\DeclareMathOperator*{\EV}{\mathbb{E}}
\newcommand{\EVV}[2][\ppvect \in \ppspace]{\EV_{#1}\left[{#2}\right]}
\newcommand{\norm}[2][\infty]{\left\|#2\right\|_{#1}}
\newcommand{\cl}{\delta}
\newcommand{\loss}{\mathcal{J}}
\newcommand{\lossf}[2][]{\mathcal{J}#1\left(#2\right)}
\newcommand{\pdim}{d}
\newcommand{\plb}{\varUpsilon}

\author{
  Matteo Pirotta \\
  INRIA Lille - Sequel Team\\
  Politecnico di Milano, Milan, Italy\\
  \texttt{matteo.pirotta@inria.fr} \\
  \And
  Marcello Restelli\\
  Dept. of Elec., Info. and Bioeng.\\
  Politecnico di Milano, Milan, Italy\\
  \texttt{marcello.restelli@polimi.it} \\
}

\title{Cost-Sensitive Approach to Batch Size Adaptation\\ for Gradient Descent}

\hypersetup{
pdfinfo={
    Title={Cost-Sensitive Approach to Batch Size Adaptation for Gradient Descent},
    Author={Matteo Pirotta, Marcello Restelli}
  },
pdfproducer={}
}

\begin{document}

\maketitle

\begin{abstract}
In this paper we propose a novel approach to automatically determine the batch size in stochastic gradient descent methods.
The choice of the batch size induces a trade-off between the accuracy of the gradient estimate and the cost in terms of samples of each update.
We propose to determine the batch size by optimizing the ratio between a lower bound to a linear or quadratic Taylor approximation of the expected improvement and the number of samples used to estimate the gradient.
The performance of the proposed approach is empirically compared with related methods on popular classification tasks.

The work was presented at the NIPS workshop on Optimizing the Optimizers. Barcelona, Spain, 2016.
\end{abstract}

\section{Introduction}
The optimization of the expectation of a function is a relevant problem in large-scale machine learning and in many stochastic optimization problems involving finance, signal processing, neural networks, just to mention a few.
The availability of large datasets has called the attention on algorithms that scale favorably both with the number of trainable parameters and the size of the input data.
Batch approaches that exploit a large number of samples to compute an approximation of the gradient have been gradually replaced by stochastic approaches that sample a small dataset (usually a single point) per iteration.
For example, \emph{stochastic gradient descent} (SGD) methods have been observed to yield faster convergence and (sometimes) lower test errors than standard batch methods~\citep{bottou2011tradeoffs}.

Despite the optimality results~\citep{bottou2004large} and the successful applications, in practice SGD requires several steps of manual adjustment of the parameters to obtain good performance.
For example, the initial step size together with the design of an appropriate annealing schema is required for learning with stationary data~\citep{bottou2012tricks,schaul2013nolearningrates}.
In addition, to limit the effects of noisy updates, it is often necessary to exploit mini-batch techniques that require the choice of an additional parameter: the \textit{batch size}.
This optimization is costly and tedious since parameters have to be tested on several iterations.
Such problems get even worse when nonstationary settings are considered~\citep{schaul2013nolearningrates}.

Several techniques have been designed for the tuning of the step size with pure SGD method.
Although these approaches have been successful applied to mini-batch settings, the design of the appropriate batch size is still an open problem.
The contribute of this paper is the derivation of a novel algorithm for the selection of the batch size in order to compromise between noisy updates and more certain but expensive steps.
The proposed algorithm \emph{automatically} adapts the batch size at each iteration in order to maximize a lower bound to the expected improvement by accounting for the cost of processing samples.
In particular, we consider both a first-order and a second-order Taylor approximation of the expected improvement and, exploiting concentration inequalities, we compute lower bounds to such approximations.
The batch size is chosen by maximizing the ratio between the lower bound to the expected improvement and the number of samples used to estimate the gradient.
Such optimization problem trades off the desire of increasing the batch size to get more accurate estimates and the cost of using more samples.
The only parameter to be handled is the probability $\cl$ that regulates the confidence level of the lower bound to the improvement step.

The rest of the paper is organized as follows.
In the next section we give a brief overview of stochastic gradient descent methods.
In Section~\ref{S:CostSensitiveScenario} we define the optimization problem used to select the batch size.
Section~\ref{S:LowerBound} introduces an approximation of the expected improvement expoliting Taylor expansion and Sections~\ref{S:L-PAST} and~\ref{S:Q-PAST} deals respectively with a linear and a quadratic Taylor approximation. Section~\ref{S.DiagonalPreconditioning} discuss the application of diagonal preconditioning to define dimension-dependent step sizes.
Empirical comparisons of the proposed methods with related approaches are reported in Section~\ref{S:Experiments}, while Section~\ref{S:Conclusions} draws conclusions and outlines future work.

%
%
%

\section{Background}\label{S:Backgorund}

Stochastic gradient descent (SGD) is one of the most important optimization methods in machine learning.
Most of the research on SGD has focused on the choice of the step size~\citep{peters2007machine,roux2010fast,duchi2011adagrad,zeiler2012adadelta,schaul2013nolearningrates,orabona2014simultaneous}.
Several annealing schemes have been proposed in literature based on the standard rule $\lstepf{t} = \lstep_0\left(1+\gamma t \right)^{-1}$ originally proposed in~\citep{robbins1951stochastic} and analyzed in~\citep{xu2011optimal,bach2011saa}.
More recently researchers have proposed techniques to adapt the step size online accordingly to the observed samples and gradients.
These techniques derive a global step size or adapt the step size for each parameter (\emph{diagonal preconditioning}).
Refer to~\citep{george2006adaptive} for a survey on annealing schemes and adaptive step size rules.

Traditional SGD processes one example per iteration. This sequential nature makes SGD challenging for distributed inference.
A common practical solution is to employ minibatch training, which aggregates multiple examples at each iteration.
On the other hand, the choice of the batch size is critical since too small batches lead to high communication costs, while large batches may slow down convergence rate in practice~\citep{li2014efficient}.
Despite the increasing amount of research in this field, all the mentioned approaches focus on obtaining (sub)optimal convergence rate of SGD without considering the possibility to adapt the size of the mini-batch.
A notable exception is the work presented in~\citep{byrd2012sample} where the authors proposed to adapt the sample size during the algorithm progression.
The batch size is selected according to the variance of the gradient estimated from observed samples.
Starting from the geometrical definition of descent direction, through several manipulations, the authors derived the following condition
\begin{equation}
\label{E:adss_rule}
|S| \geq \frac{\norm[1]{\text{Var}\left[{\nabla_{\vtheta}}\right]}}{\gamma^2 \norm[2]{\nabla_{\vtheta}^S}^2},
\end{equation}
where $\gamma \in (0,1)$ and $\text{Var}\left[\nabla_{\vtheta}\right]$ is the vector storing the population variance for each component ($\text{Var}\left[{\nabla_{\vtheta}}\right] = \transpose{[\sigma^2_1,\ldots, \sigma^2_{\pdim}]}$).
The population variance is then approximated through its unbiased estimate $\text{Var}\left[\nabla_{\vtheta}^S\right]$ computed on sample set $S$.
However, the interplay between sample size and step size is not investigated resulting in an algorithm with hyper-parameters for both the selection of the batch size (the meaning of $\gamma$ is not clearly defined) and the tuning of the step size.

\section{Cost Sensitive Scenario}\label{S:CostSensitiveScenario}
In this section we formalize the problem and the methodology that will be used through all the paper.
Consider the problem of maximizing the expected value of a function $f$ (we assume that $f$ is Lipschitz continuous with Lipschitz constant $L$):
$$\max_{\vtheta} \lossf{\vtheta} = \max_{\vtheta} \EVV[x \sim \mathcal{P}]{f(x,\vtheta)},$$
where $\vtheta \in \realspace^\pdim$ is the trainable parameter vector and the samples $x$ are drawn i.i.d. from a distribution $\mathcal{P}$.
A common approach is to optimize the previous function through gradient ascent. 
However, since $\mathcal{P}$ is unknown it is not possible to compute the \emph{exact} gradient $\nabla_{\vtheta}\loss$, but we can estimate it through samples.
Given a training set $S = \left\{ x_i | x_i \sim \mathcal P, i=1,\ldots, N\right\}$, the mini-batch stochastic gradient (SG) ascent is a stochastic process
\begin{equation}
\label{E:updateSG}
\begin{aligned}
\vtheta^{(t+1)} 
&= \vtheta^{(t)} + \Delta \vtheta^n\\
&= \vtheta^{(t)} + \vlstepf{t} \nabla_{\vtheta} \loss^n, \qquad t \in \mathbb{N}^+
\end{aligned}
\end{equation}
where $\Delta \vtheta^n$ is random variable associated to a $n$-dimensional subset of $S$ (\eg randomly drawn).
Formally $\Delta \vtheta^n$ is defined as the product of a positive scalar (or positive semi-definite matrix) $\lstepf{t}$ and a the gradient \emph{estimate} built on a $n$ samples drawn from $S$
\begin{equation*}
 \nabla_{\vtheta} \loss^n = \frac{1}{n}\sum_{i \in \mathcal{I}^n} \nabla_{\vtheta} f(x_i, \vtheta),
\end{equation*}
where $\mathcal{I}^n$ is an index set used to identify elements in $S$.
$\nabla_{\vtheta} \loss^n$ is a random variable that depends on the selection of the subset of $S$, \ie the index set $\mathcal{I}^n$.
In the following we will show how to select the batch size $n$ for each gradient update.

To evaluate the quality of an update we consider the improvement $\imp^n = \lossf{\vtheta + \Delta \vtheta^n} - \lossf{\vtheta}$ that is again a random variable.
As the number of samples $n$ increases, the gradient estimate (and consequently the estimated improvement) gets more and more certain.
So, adopting a risk-averse approach, we consider as a goal the maximization of some statistical lower bound $\plb^n$ to the expected improvement $\imp$. 
This allows to account for the uncertainty in the stochastic process defined in~\ref{E:updateSG}.
On the other hand, this problem is trivially solved by taking the batch size as large as possible, thus not considering the additional computational cost of processing a larger batch size.
In practice, this means that the batch dimension $n$ induces a trade-off between a secure but costly update (the estimate converges to the true value as $n \to +\infty$) and a noisy one.
In order to formalize the trade-off, in this paper we consider that any additional sample comes at a price and when the addition of a new sample does not provide any significant improvement in the estimated performance it is not worth to pay that price.
As a consequence, we can formalize the batch size selection problem as a \emph{cost sensitive optimization}
\begin{equation}
\label{E:costproblem}
n^* = \argmax_{n \in \mathbb{N}^+} \frac{\plb^n}{n}.
\end{equation}

\section{Lower Bound to the Improvement}\label{S:LowerBound}
This section focuses on the derivation of the lower bound to the improvement $\Delta \loss^n$.
Given an increment $\Delta \vtheta^n$, a realization of the random variable $\Delta \loss^n$ can be computed.
However, we do not know the analytical relationship that ties the two terms.
The lack of this information prevents a closed-form solution for the optimal batch size. 
On the other hand, resorting to black-box optimization methods (\eg grid search) is generally not a suitable alternative due to their high cost. 

In order to simplify the optimization problem, we can consider the Taylor expansion of the expected improvement.
For example, the first order expansion is given by
\begin{align}
\label{E:taylor1}
\eimp
&= \transpose{\nabla_{\vtheta} \loss} \Delta \vtheta^n + R_1\left( \Delta\vtheta^n \right),
\end{align}
where $R_1\left(\Delta\vtheta^n\right)$ is the remainder.
A lower bound to the remainder is easily derived by minimizing the remainder along the line connecting the current parameterization $\vtheta$ and the value $\vtheta+\Delta\vtheta$: $R_1(\Delta \vtheta) \geq \frac{1}{2}\inf_{c \in (0,1)} \left(\transpose{\Delta\vtheta} H_{\vtheta} \lossf{\vtheta + c\, \Delta\vtheta} \Delta\vtheta \right)$.
By plugging in this result in~\eqref{E:taylor1}, a \emph{deterministic} lower bound to the improvement is derived.

This formalization has two issues. 
First, the computation of such lower bound needs to solve a minimization problem that requires the evaluation of the Hessian in several points along the gradient direction.
Secondly, the above lower bound does not explicitly depend on the batch size, since it does not take into consideration the uncertainty in the gradient estimate.
The first issue will be solved by considering an approximation of the expected improvement obtained by considering a truncation of the Taylor expansion, while the second issue is addressed by considering a probabilistic lower bound to the expected improvement, that explicitly depends on the batch size (uncertainty reduces as batch size increases).


\subsection{Approximation of the Expected Improvement}
\label{S:simplified}
As mentioned, the computation of the lower bound to the remainder of the first-order Taylor expansion requires the evaluation of the Hessian in several points (depending on $c$) which has a quadratic cost in the number of parameters $\pdim$.
One way to deal with this issue is to require a high order Lipschitz continuous condition on the objective function in order to derive a bound to the Hessian or to exploit the knowledge of the objective function~\citep{pirotta2013adaptive}.
However, in practice this information is hard to retrieve and since our goal is to derive a practical algorithm we suggest to exploit approximations of the improvement $\eimp$.

A first formulation is obtained by considering a local \emph{linear expansion}: $\eimp \approx \transpose{\nabla_{\vtheta} \loss} \Delta\vtheta^n$.
As we will see in the next section, this simplification has several advantages.
As second option, in Section~\ref{S:Q-PAST}, we suggest to replace the inferior with the evaluation of the Hessian in the current parametrization.
Equivalently, this means that we consider a \emph{quadratic expansion}, a choice that is common in literature~\citep{roux2010fast,schaul2013nolearningrates}.
Formally, we consider that
$\eimp \approx \transpose{\nabla_{\vtheta}\loss} \Delta\vtheta^n + \frac{1}{2} \transpose{\Delta\vtheta^n} H_{\vtheta}\loss \Delta\vtheta^n$, where the second-order remainder has been dropped.

\section{Linear Probabilistic Adaptive Sample Technique (L-PAST)}
\label{S:L-PAST}
The linear expansion allows to select the batch size in a way that is \emph{complementary} to the step size selection technique since it is independent from the selected step size.
In other words, the advantage of this approach is that the step size can be tuned using any automatic technique provided in literature, while the batch size is selected automatically according to the quality of the observed samples.

Let $\simpimp = \transpose{\nabla_{\vtheta} \loss} \Delta\vtheta^n$ be the linear simplification of the expect improvement.
We still need to manipulate such formulation in order to remove the dependence on the true gradient.
This goal can be achieve by exploiting concentration inequalities on the exact gradient $\nabla_{\vtheta}\loss$.
Formally, we consider that the following inequality holds with probability (\wp) $1 - \delta$
\begin{equation}
\label{E:concineq}
\norm[2]{\nabla_{\vtheta}\loss - \nabla_{\vtheta}\loss^n} < B_{\norm[]{\nabla}}(n,\delta).
\end{equation}
Given the previous inequality it is easy to prove the following bound, \wp $1 - \delta$, (see the appendix, Sec.~\ref{S:lb_sample_improvement})
\begin{align}
\label{E:eimp_bound}
\simpimp
& = \lstep\, \transpose{\nabla_{\vtheta} \loss}\, \nabla_{\vtheta} \loss^n\\
\nonumber
& > \lstep\, \left(\norm[2]{\nabla_{\vtheta}\loss^n} - B_{\norm[]{\nabla}}(n,\delta) \right) \norm[2]{\nabla_{\vtheta} \loss^n} = \plb^{n,\delta}_{L},
\end{align} 
where we have considered the global step size ($\lstep \in \realspace^+$).
As expected, the lower bound to the expected improvement depends on the batch size through the concentration bound ($\norm[2]{\nabla_{\vtheta}\loss^n}$ is a realization of the random variable given the current set $\mathcal{I}^n$).
In particular, as the number of samples increases, the empirical error (according to the concentration inequality) decreases leading to better estimates of the expected improvement.
Having derived a sample-based bound to the expected improvement, we can solve the cost-sensitive problem~\eqref{E:costproblem} for the ``optimal'' batch size $n$.
L-PAST is outlined in Algorithm~\ref{A:lpast}.

\begin{algorithm}[t]
\caption{L-PAST}
\label{A:lpast}
\begin{algorithmic} 
\STATE \textbf{Inputs:} Sample set $S=\{x_1,\dots,x_N\}$, initial batch size $n$, confidence level $\delta$
\FOR{t=1 \TO T}
\STATE $\mathcal{I}^n \leftarrow \left\{i | i \in {1,\ldots, N} \wedge |\mathcal{I}^n| = n\right\}$
\STATE $\nabla_{\vtheta} \loss^n \leftarrow \frac{1}{n}\sum_{i \in \mathcal{I}^n} \nabla_{\vtheta} f\left(x_i, \vtheta^{(t)}\right)$
\STATE $\vtheta^{(t+1)} \leftarrow \vtheta^{(t)} + \lstep \nabla_{\vtheta} \loss^n$ 
\STATE $n \leftarrow \argmax_{n \in \mathbb{N}^+} \frac{\plb_{L}^{n,\delta}}{n}$
\ENDFOR
\end{algorithmic}
\end{algorithm}

\begin{table*}[t]
\centering
\begin{tabular}{ccc}
\hline
Hoeffding & Chebyshev & Bernstein\\
\hline
\rule{0pt}{3.5ex}
$n \geq \frac{18 L^2}{\norm[2]{\nabla_{\vtheta}\loss^n}^2} \ln\left(
\frac{\pdim+1}{\delta}
\right)$&
$n \geq \frac{9 \norm[1]{\text{Var}[\nabla_{\vtheta}\loss]}}{4\delta\norm[2]{\nabla_{\vtheta}\loss^n}^2}$&
$ n \geq \frac{9b+16a\norm[2]{\nabla_{\vtheta} \loss^n}+3\sqrt{9b^2+32ab \norm[2]{\nabla_{\vtheta} \loss^n}}}{8\norm[2]{\nabla_{\vtheta} \loss^n}^2}$\\[.2cm]
\hline
\end{tabular}
\caption{Batch size obtained by solving Problem~\eqref{E:costproblem} using different concentration inequalities. We used the symbols $a$ and $b$ to simplify the Bernstein's size ($a = \frac{2}{3} L \ln\left(\frac{d+1}{\delta}\right)$, $b = 2 \norm[2]{\text{Var}[\nabla_{\vtheta}\loss]}\ln\left(\frac{d+1}{\delta}\right)$).}
\label{T:batchdim}
\end{table*}
\subsection{Concentration Inequalities and Batch Size}
\label{S:linear_concentration}
The bound in~\eqref{E:eimp_bound} provides a generic lower bound to the expected improvement that is independent from the specific concentration inequality that is used. It is now necessary to provide an explicit formulation in order to solve Problem~\eqref{E:costproblem}.
Several concentration inequalities have been provided in literature, in this paper we consider Hoeffding's, Chebyshev's and Bernstein's inequalities~\citep{massart2007concentration}.
Chebyshev's inequality has been widely exploited in literature due to its simplicity, it can be applied to any arbitrary distribution (by knowing the variance).
On the other side, Hoeffding's and Bernstein's inequalities require a bounded support of the distribution, \ie the knowledge of the range of the random variables (here $\nabla_{\vtheta} f(x_i, \vtheta)$).
We use the term \emph{distribution aware} to refer to the scenario where the properties of the distribution are known (\eg variance and range).
Although these values can be \emph{estimated} online from the observed value, the results may be unreliable in the event of poor estimates.
Empirical versions---that directly account for the estimation error---have been presented in literature~\citep{saw1984chebyshev,mnih2008empirical,stellato2016multivariate}.

The advantage of using these inequalities is that the batch size can be easily computed in \emph{closed form}, see Table~\ref{T:batchdim} for the distribution aware scenario.
It is worth to notice that all the proposed approaches retains one hyper-parameter $\delta\in (0,1)$ which denotes the desired confidence level.
This parameter can be easily set due to its clear meaning and typically its contribution is small since it is inside a logarithm.

It is worth to notice that, when we consider the Chebyshev's inequality, \ie $B_{_{\norm[]{\nabla}}}^{n,\delta} = \sqrt{\frac{\norm[1]{\text{Var}\left[\nabla_{\vtheta}\loss\right]}}{n\delta}}$, our approach provides a probabilistic interpretation of the AGSS algorithm presented in~\citep{byrd2012sample} and reported in~\eqref{E:adss_rule}.
Nevertheless, our derivation gives a different and more formal interpretation of their approach and gives an explicit meaning to the hyper-parameter by mapping $\frac{4\delta}{9}$ to $\gamma^2$.
It is worth to notice that this result is obtained by considering the distribution aware Chebyshev's inequality instead of the empirical version.
By replacing the variance with its empirical estimate the result may be unreliable.

\section{Quadratic Probabilistic Adaptive Sample Technique (Q-PAST)}\label{S:Q-PAST}
The simplicity of the previous approach comes at a low expressive power.
The quadratic expansion of the expected improvement (with \emph{global} step size)
\begin{align}
\label{E:qsimpimp}
\simpimp 
&= \transpose{\nabla_{\vtheta}\loss} \Delta\vtheta^n + \frac{1}{2} \transpose{\Delta\vtheta^n} H_{\vtheta}\loss \Delta\vtheta^n\\
&= 
\lstep\, \transpose{\nabla_{\vtheta}\loss} \nabla_{\vtheta} \loss^n +
\frac{1}{2}\lstep^2 \transpose{\left( \nabla_{\vtheta} \loss^n \right)}
H_{\vtheta}\loss
\nabla_{\vtheta} \loss^n \nonumber
\end{align}
allows to account for local curvatures of the space.

Before to describe the Quadratic-PAST (Q-PAST), as done with L-PAST, we need to manipulate the expected improvement in order to remove the dependence on the exact gradient and Hessian.
While the linear term can be lower bounded as done in Section~\ref{S:L-PAST}, here we show how to handle the quadratic form in a similar way.
Consider a component-wise concentration inequality for the Hessian estimate, such that, \wp $1-\delta/(2d^2)$:
\begin{equation}
 \left| H_{\vtheta}^{(ij)} \loss - H_{\vtheta}^{(ij)} \loss^n \right| < B_H^{(ij)}\left(n,\frac{\delta}{2d^2}\right) \quad \forall i,j.
\end{equation}
Then,
\begin{align}
\transpose{\nabla_{\vtheta}\loss^n}\; H_{\vtheta}\loss \; \nabla_{\vtheta} \loss^n
\label{E:lbquadratic}
>
\transpose{\nabla_{\vtheta}\loss^n}\; \widetilde{H}_{\vtheta}\loss^n \; \nabla_{\vtheta} \loss^n,
\end{align}
where
$\widetilde{H}_{\vtheta}^{(ij)}\loss^n = H_{\vtheta}^{(ij)}\loss^n - B_H^{(ij)}(n,\delta)$.
By plugging in inequalities~\eqref{E:eimp_bound}--\eqref{E:lbquadratic} in~\eqref{E:qsimpimp} we obtain a lower bound to the quadratic expansion of the improvement
\begin{align}
\label{E:lbq}
 \simpimp &> \plb_{L}^{n,\frac{\delta}{2}} + \frac{1}{2}\lstep^2 \transpose{\nabla_{\vtheta}\loss^n}\; \widetilde{H}_{\vtheta}\loss^n \; \nabla_{\vtheta} \loss^n \\
 \notag
 &= \plb_{Q}^{n,\delta}.
\end{align}

Given the step size $\lstep$ and a set $\mathcal{I}^n$, we can optimize the lower bound for the batch size $n$.

Finally, we can exploit this sample-based bound to compute the ``optimal'' \emph{batch size} as in Problem~\eqref{E:costproblem}.
The concentration inequalities mentioned in Section~\ref{S:linear_concentration} can be used to bound the Hessian components.
By exploiting these bounds it is possible to derive closed--form solution for $n$ even in this context.

\begin{algorithm}[t]
\caption{Q-PAST}
\label{A:qpast}
\begin{algorithmic} 
\STATE \textbf{Inputs:} Sample set $S=\{x_1,\dots,x_N\}$, initial batch size $n$, confidence level $\delta$
\FOR{t=1 \TO T}
\STATE $\mathcal{I}^n \leftarrow \left\{i | i \in {1,\ldots, N} \wedge |\mathcal{I}^n| = n\right\}$
\STATE $\nabla_{\vtheta} \loss^n \leftarrow \frac{1}{n}\sum_{i \in \mathcal{I}^n} \nabla_{\vtheta} f\left(x_i, \vtheta^{(t)}\right)$
\STATE $\vtheta^{(t+1)} \leftarrow \vtheta^{(t)} + \lstep \nabla_{\vtheta} \loss^n$ 
\STATE $H_{\vtheta} \loss^n \leftarrow \frac{1}{n}\sum_{i \in \mathcal{I}^n} H_{\vtheta} f\left(x_i, \vtheta^{(t)}\right)$
\STATE $n \leftarrow \argmax_{n \in \mathbb{N}^+} \frac{\plb_{Q}^{n,\delta}}{n}$
\ENDFOR
\end{algorithmic}
\end{algorithm}

\section{Diagonal Preconditioning}\label{S.DiagonalPreconditioning}
Until now we have considered the global step size scenario where each parameter is scaled by the same amount $\lstep$.
In practice, it may be necessary to considered individual step sizes ($\Delta\vtheta_i = \lstep_i\nabla_{\vtheta}\loss^n_i$) in order to account for the different magnitudes of the parameters.

There are several ways to deal with such scenario.
We start considering the linear expansion $\simpimp = \transpose{\left( \vlstep^{\frac{1}{2}} \circ \nabla_{\vtheta} \loss \right)} \left( \vlstep^{\frac{1}{2}} \circ \nabla_{\vtheta} \loss^n \right)$
 where $\vlstep$ is a $\pdim$-dimensional vector and $\circ$ is the Hadamard (element-wise) product.
If now we consider the gradient to be scaled by a factor $\vlstep^{\frac{1}{2}}$, we can apply the same procedure presented in Section~\ref{S:L-PAST} on the scaled gradient.
This means that it is necessary to recompute the concentration inequalities to take into account the change of magnitude.
For example, Hoeffding's inequality requires to know an upper bound to the L2--norm of the random vector involved in the estimate.
In our settings (see Section~\ref{S:CostSensitiveScenario}) we have assumed that $\norm[2]{\nabla_{\vtheta} f(x_k, \vtheta)} \leq L$ for any $x_k$ and $\vtheta$.
To use a diagonal preconditioning with L-PAST and Hoeffding we need just compute the upper bound to $\norm[2]{\vlstep \circ \nabla_{\vtheta} f(x_k, \vtheta)}$ that in a trivial form is:
$\norm[2]{\vlstep \circ \nabla_{\vtheta} f(x_k, \vtheta)} \leq L \max_i \vlstep_i $.
Similar considerations can be derived for the other concentration inequalities.

Another possible way to deal with the diagonal preconditioning is to exploit a component-wise concentration inequality.
Let $\boldsymbol{B}_{\nabla}$ be a vector such that
$\left| \nabla_{\vtheta} \loss_i - \nabla_{\vtheta} \loss^n_i \right| < \boldsymbol{B}_{\nabla}^{(i)}$ \wp $1-\delta/2d$.
This is the element-wise counterpart of the concentration inequality considered in~\eqref{E:concineq}.
Notice that the following inequality always holds: $B_{\norm[]{\nabla}} \leq \pdim \max_i \boldsymbol{B}_{\nabla}^{(i)}$.
Let us consider this scenario together with the quadratic expansion, then:
\begin{align}
\label{E:imprdiag}
\simpimp
&\geq
\transpose{
\left( \vlstep \circ \nabla_{\vtheta}\loss^n \right)
}
\;
\widetilde{\nabla}_{\vtheta} \loss^n\\
\nonumber
&+ \frac{1}{2} \transpose{
\left( \vlstep \circ \nabla_{\vtheta}\loss^n \right)
}\;
\widetilde{H}_{\vtheta} \loss^n
\;
\left( \vlstep \circ \nabla_{\vtheta}\loss^n \right) \; \text{\wp }\; 1-\delta
\end{align}
where $\widetilde{\nabla} \loss^n = \nabla_{\vtheta} \loss^n - \boldsymbol{B}_{\nabla}$ and
$\widetilde{H}_{\vtheta} \loss^n$ is defined as in~\eqref{E:lbquadratic}.

A different way to deal with diagonal preconditioning is to assume the problem to be separable~\citep{schaul2013nolearningrates}.
In our settings this maps to a diagonal approximation of the Hessian $\widetilde{H}_{\vtheta} \loss^n$.

\section{Experiments}\label{S:Experiments}
We tested the approaches on digit recognition and news classification tasks, with both convex (logistic regression) and non-convex (multi-layer perceptron) models.

Mini-batch SG (SG-$n$) with fixed batch size $n$ is the standard approach to stochastic optimization problems.
This section compares SG-$n$ with the adaptive algorithms introduced above (three versions of PAST and DSG~\citep{byrd2012sample}).

A critical parameter in SG optimization is the definition of the step length $\lstep$.
In order to remove the dependence from these parameters we have tested several adaptive strategies (\eg AdaGrad, Adam, RMSprop, AdaDelta).
We have finally decided to use RMSprop which provided the most consistent results across the different settings (parameters are set as suggested in~\citep{tielemanH12}).
Finally, the $n$-dimensional subset of $S$ is sampled sequentially without shuffling at the beginning of each epoch.

\paragraph{Evaluation.}
The main measure to be considered is the loss $\lossf{\vtheta}$.
However, the evaluation needs to take into account two orthogonal dimensions: \emph{samples} and \emph{iterations}.
The number of samples processed by the algorithm is relevant in applications where the samples need to be actively collected.
For example, it is a relevant measure in reinforcement learning problems where samples are obtained interacting with a real or simulated environment.
On contrary, in off-line applications (\eg supervise learning) the iterations play a central role because there is no cost in collecting samples.
For example, the highest cost in deep learning approaches is the evaluation of the evaluation the computation of the gradient and the consequent update of the parameters.
Clearly this cost is proportional to the number of iterations.
In the following we will investigate both the dimensions.

\paragraph{Datasets.}

We chose to test the algorithms both classification and regression tasks.
The classification tasks are: the MNIST digit recognition task~\citep{lecun1998gradient} (with $60k$ training samples, $10k$ test samples, and $10$ classes), and a subset of the Reuters newswire topics classification\footnote{Reuters data are available at \url{https://keras.io/datasets/}.} (with $8982$ training samples, $2246$ test samples, and $46$ classes).
For the Reuters task we select the $1000$ most frequent words and we used them as binary features.
The regression task is performed on the Parkinsons Telemonitoring dataset~\citep{tsanas2010accurate}.\footnote{The dataset is available at \url{https://archive.ics.uci.edu/ml/datasets/Parkinsons+Telemonitoring}.}
This dataset is composed by $5875$ voice measurements and the goal is to predict the \emph{total\_UPDRS} field by using the $19$ available features. 
We did not used any form of preprocessing for the classification tasks, while we performed normalization for the Parkinsons one (zero mean and unitary variance).

\paragraph{Estimators.}
The \emph{multi-class classifier} was modeled through different architectures of feed-forward neural networks.
The simplest one is a logistic regression (\ie a network without hidden layers).
This model has convex loss (categorical cross entropy) in the parameters.
This configuration is denoted as 'M0'.
The second configuration is a fully connected multi-layer perceptron with one hidden layer with RELU activation function.
In the MNIST task the network (denoted 'M1') has the following configuration ($[784, 128, 10]$), it is not used in Reuters.
Finally we test a deep, fully connected multi-layer perceptron with two hidden layers with activation function RELU.
This architecture has been used only in the MNIST problem (denoted 'M2') with the following layers ($[784, 256, 128, 10]$).
The multi-layer parceptrons have non-convex loss (cross-entropy) relative to parameters. 

For the regression task we have decided to exploit a simple \emph{linear regressor}.
We are aware of the limited power of such estimator but the focus is not on the final performance but on the relationship between the different batch strategies.

\begin{figure*}
\centering
\includegraphics[width=.31\textwidth]{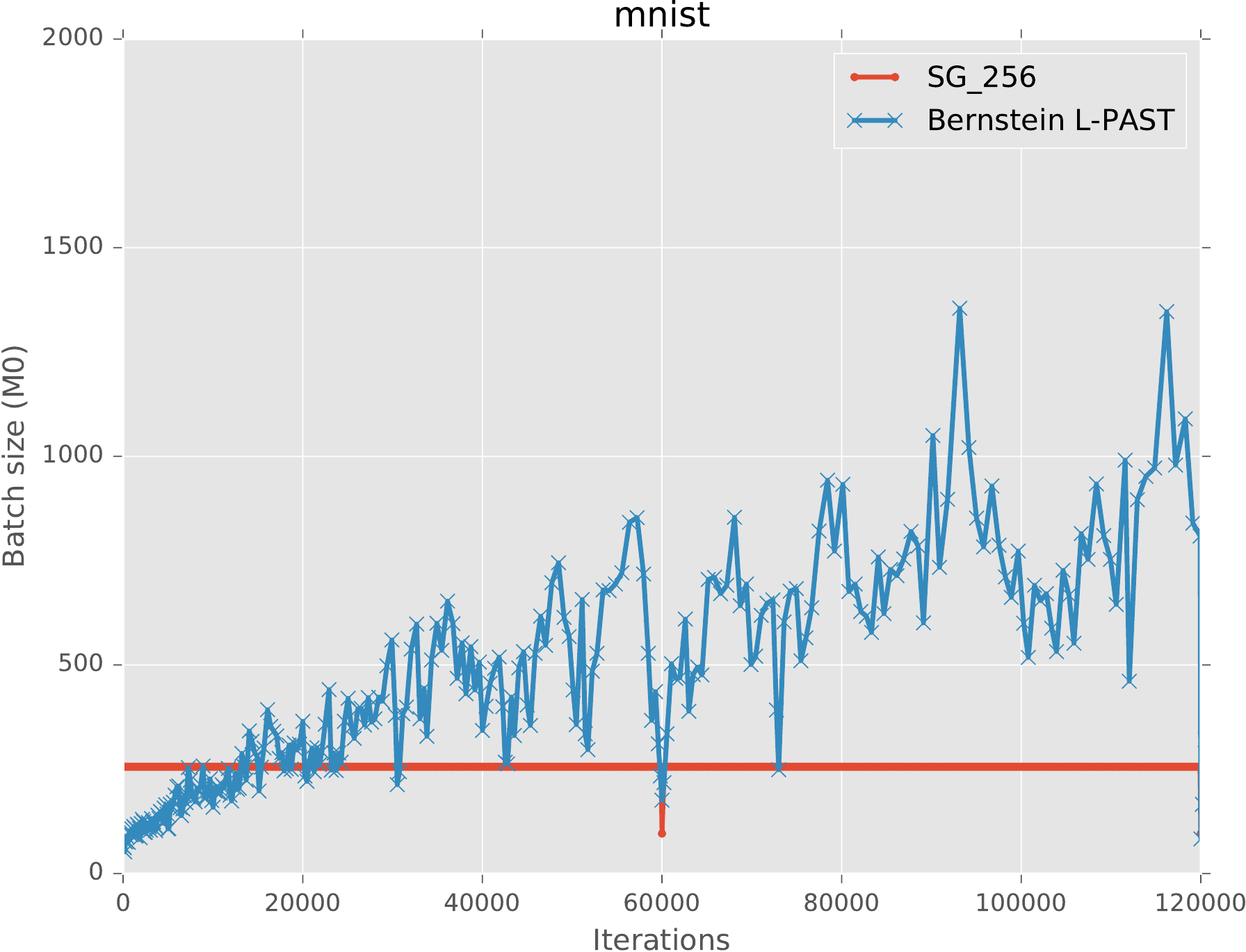}\hspace{.5cm}
\includegraphics[width=.3\textwidth]{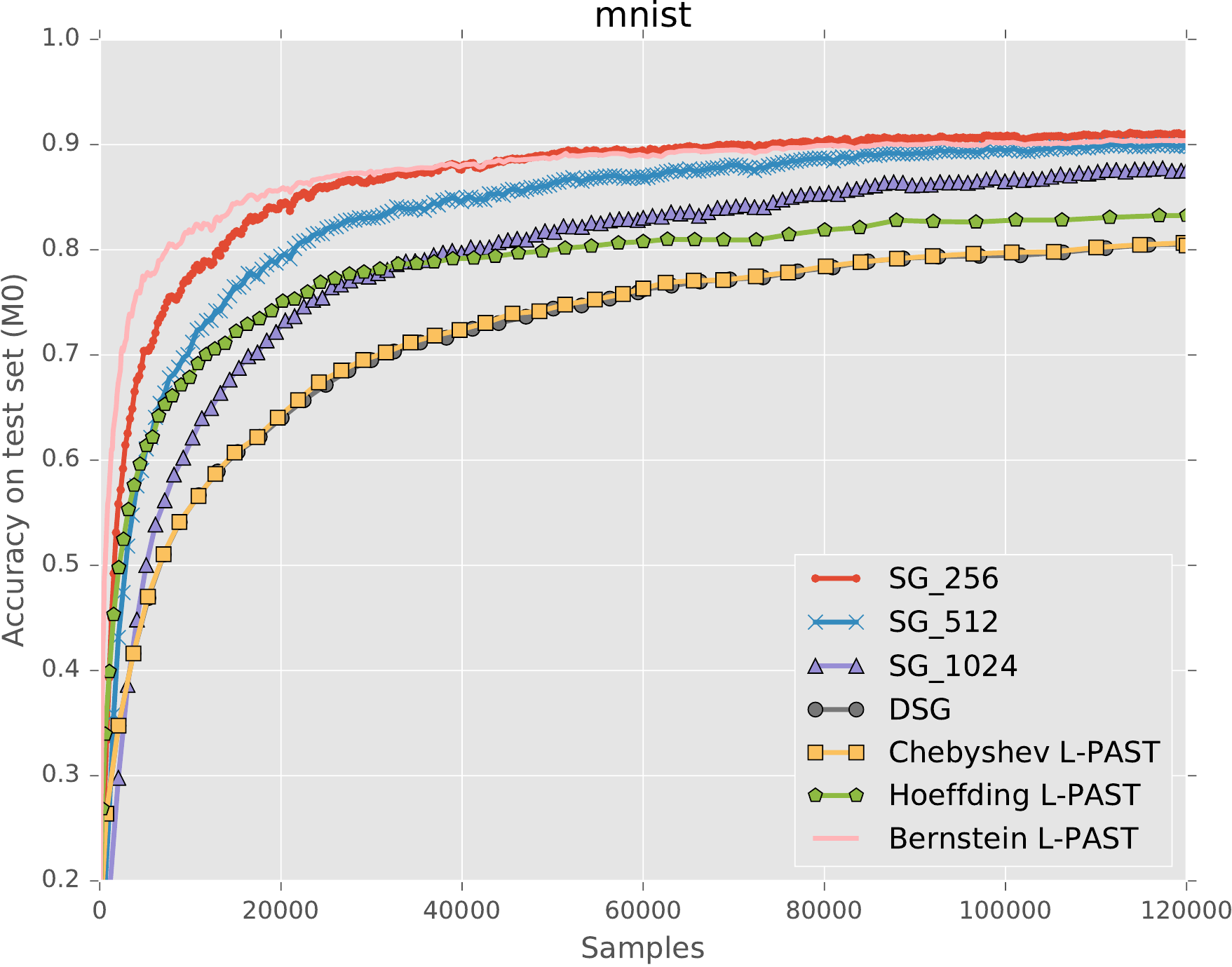}\hspace{.5cm}
\includegraphics[width=.3\textwidth]{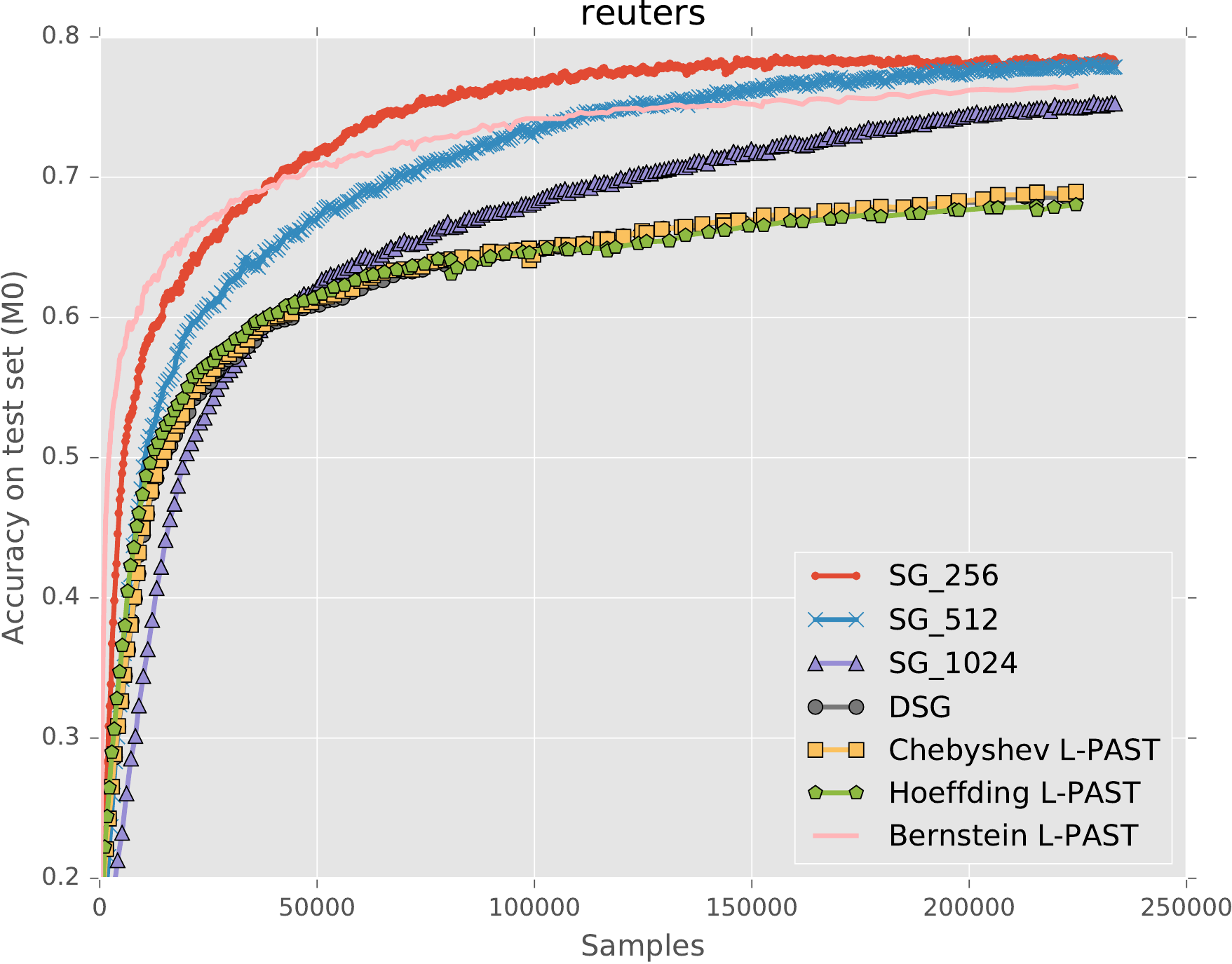}\\[.1cm]
\hspace{.31\textwidth}\hspace{.5cm}\includegraphics[width=.3\textwidth]{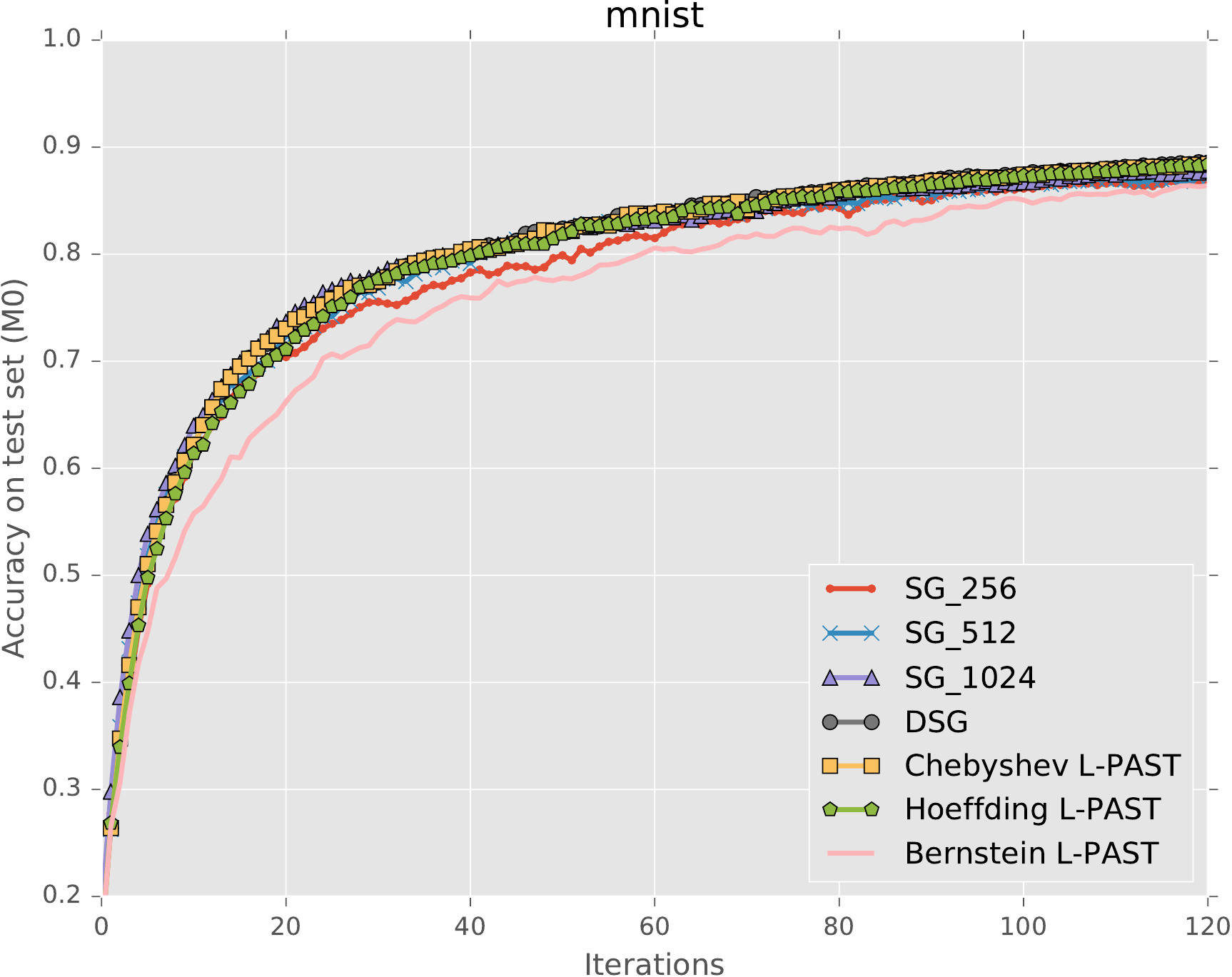}\hspace{.5cm}
\includegraphics[width=.3\textwidth]{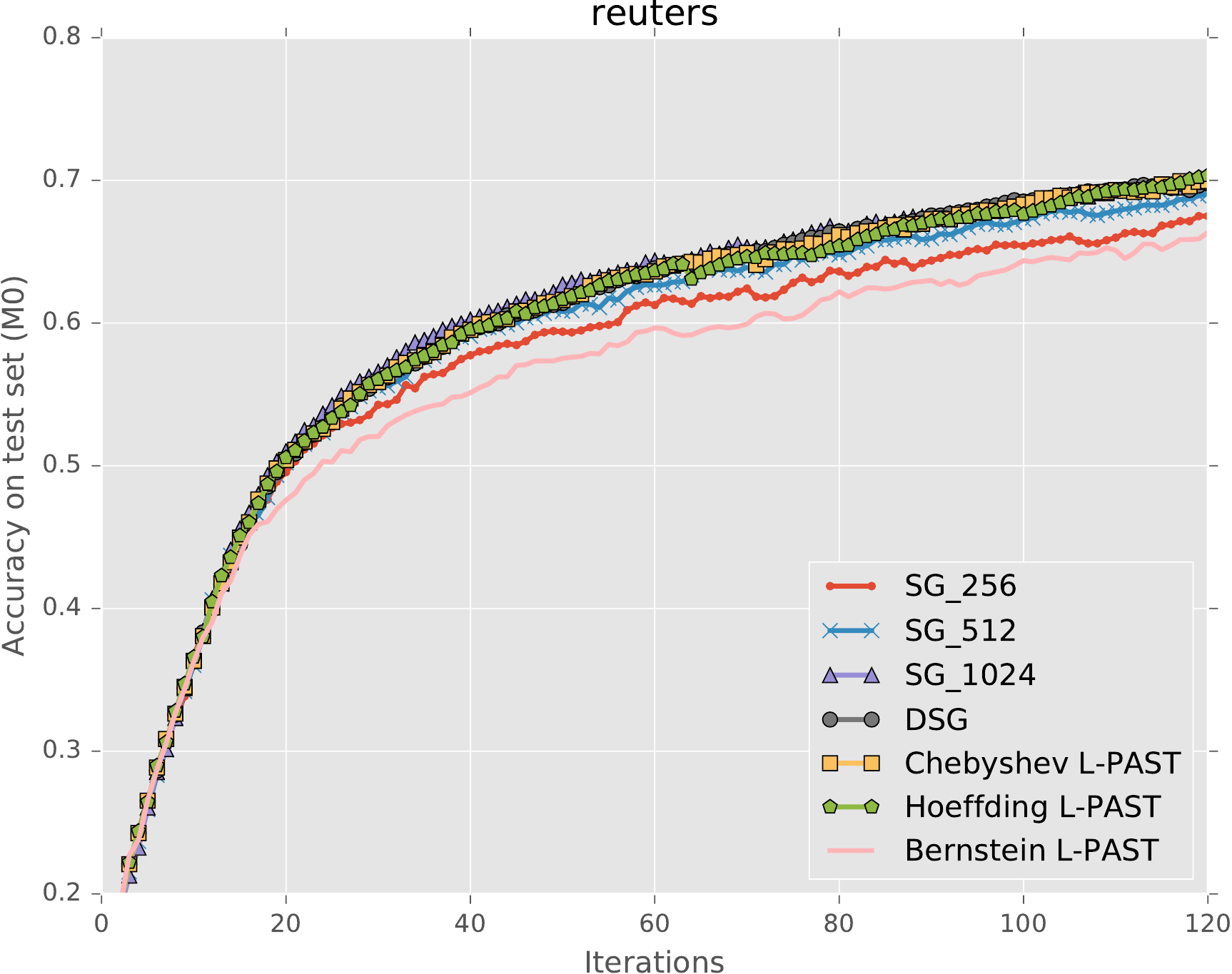}
\caption{Accuracy on test set. The net is the M0 in both the domain.}
\label{F:comparison_samples_iter}
\end{figure*}


\subsection{L-PAST Behavior.}
In this section we compare the behavior of L-PAST approaches with the state-of-the-art on the classification tasks.


We start considering the total \emph{number of processed samples} as evaluation dimension (together with the accuracy score).
As shown in Figure~\ref{F:comparison_samples_iter} (top line), the best accuracy is obtained by the algorithms that select small batches (\eg Bernstein L-PAST).
This clearly is a consequence of the higher number of updates performed by the algorithms that select small batch sizes.
In particular Bernstein L-PAST is able to outperform the other algorithms in the MNIST task due to the ability of quickly approaching the optimal solution in the initial phase.
The rightmost figure shows the number of samples selected by Bernstein L-PAST \wrt SG-$256$.
Other approaches (DSG, Hoeffding/Chebyshev L-PAST) that exploit more general inequalities are prone to select bigger batch sizes that result in less updates and slower convergence rates.

When we consider the number of iterations, the ranking of the algorithms changes. In particular the ones that select the smallest batch sizes are penalized by the noisy estimate of the gradient.
The other algorithms perform updates that are more certain and leads to highest scores.


Finally, we tested also different confidence levels. Table~\ref{T:levels} shows that, as expected, the confidence $\delta$ has a small influence on the overall behavior.
It is worth to notice that smaller batches generally lead to better (maybe noisy) performance since are able to perform a larger number of updates.

\begin{table*}[t]
 \centering
 \begin{tabular}{cccccc}
 \hline
 Model&$\delta$&DSG&H. L-PAST& C. L-PAST&B. L-PAST\\
 \hline
 M2&$0.10$&$84.93\%$ $(13)$&$89.69\%$ $(37)$&$83.37\%$ $(16)$&$\boldsymbol{96.55\%}$ $(791)$\\
 M2&$0.25$&$87.31\%$ $(25)$&$89.86\%$ $(36)$&$88.33\%$ $(27)$&$\boldsymbol{96.64\%}$ $(850)$\\
 M2&$0.50$&$90.28\%$ $(42)$&$89.25\%$ $(43)$&$91.69\%$ $(44)$&$\boldsymbol{96.63\%}$ $(902)$\\
 \hline
 M1&$0.1$&
$86.77\%$ $(21)$&
$86.95\%$ $(21)$&
$88.20\%$ $(36)$&
$\boldsymbol{94.55\%}$ $(368)$\\
M1&$0.2$&
$89.10\%$ $(31)$&
$89.09\%$ $(32)$&
$89.63\%$ $(37)$&
$\boldsymbol{94.67\%}$ $(422)$\\
M1&$0.5$&
$89.94\%$ $(46)$&
$90.75\%$ $(56)$&
$88.79\%$ $(38)$&
$\boldsymbol{94.74\%}$ $(464)$\\
\hline
 \end{tabular}
 \caption{MNIST Task. Accuracy and number of iterations with different confidence levels. SG-256, SG-512 and SG-1024 obtained
 $96.20\%$ $(705)$, $95.45\%$ $(354)$ and $94.92\%$ $(177)$ with model M2, while $95.20\%$, $94.23\%$ and $93.12\%$ with model M1. The algorithms have been trained over $3$ epochs.}
 \label{T:levels}
\end{table*}

\begin{figure}[t]
 \centering
 \includegraphics[width=.48\columnwidth]{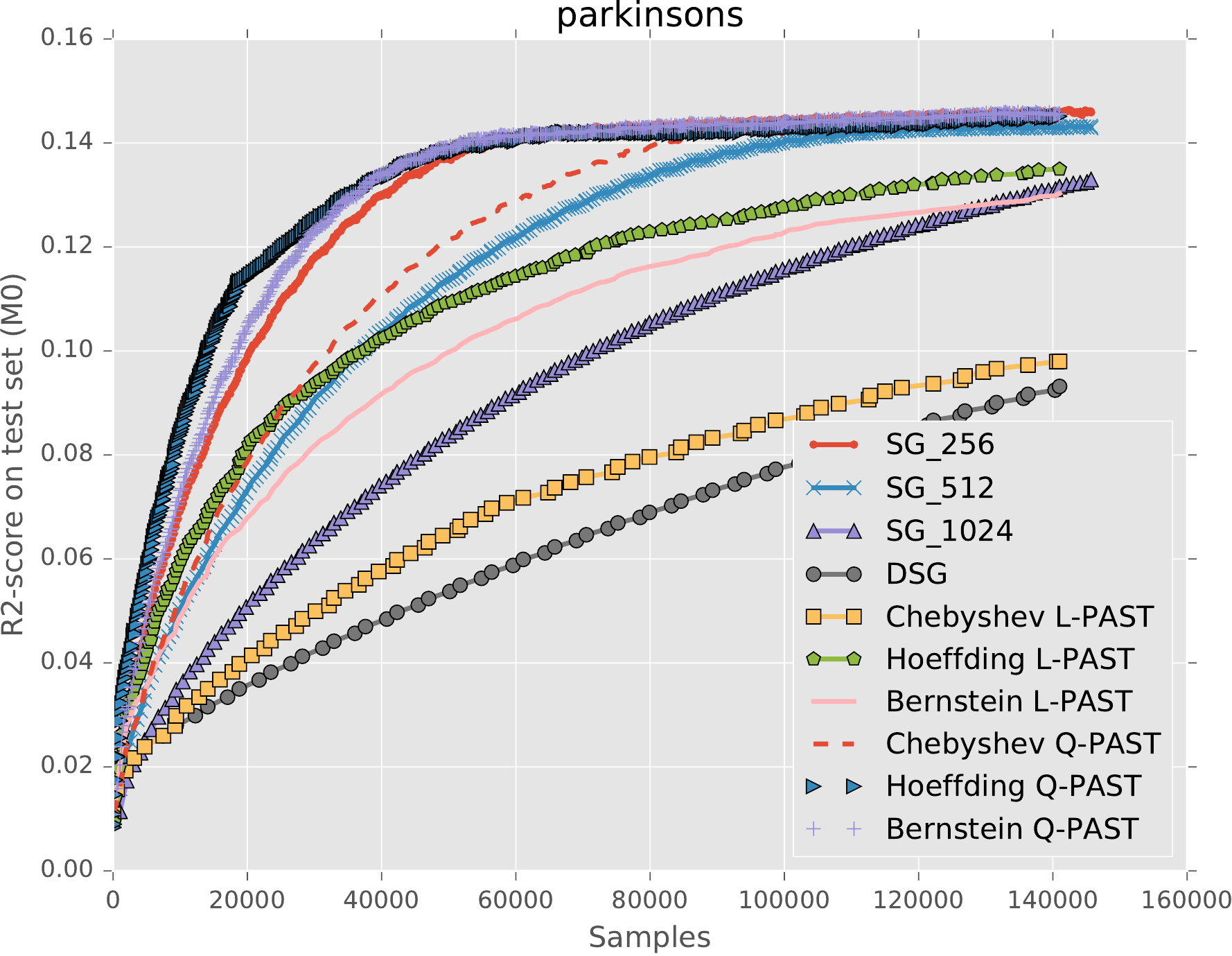}\hfill
 \includegraphics[width=.48\columnwidth]{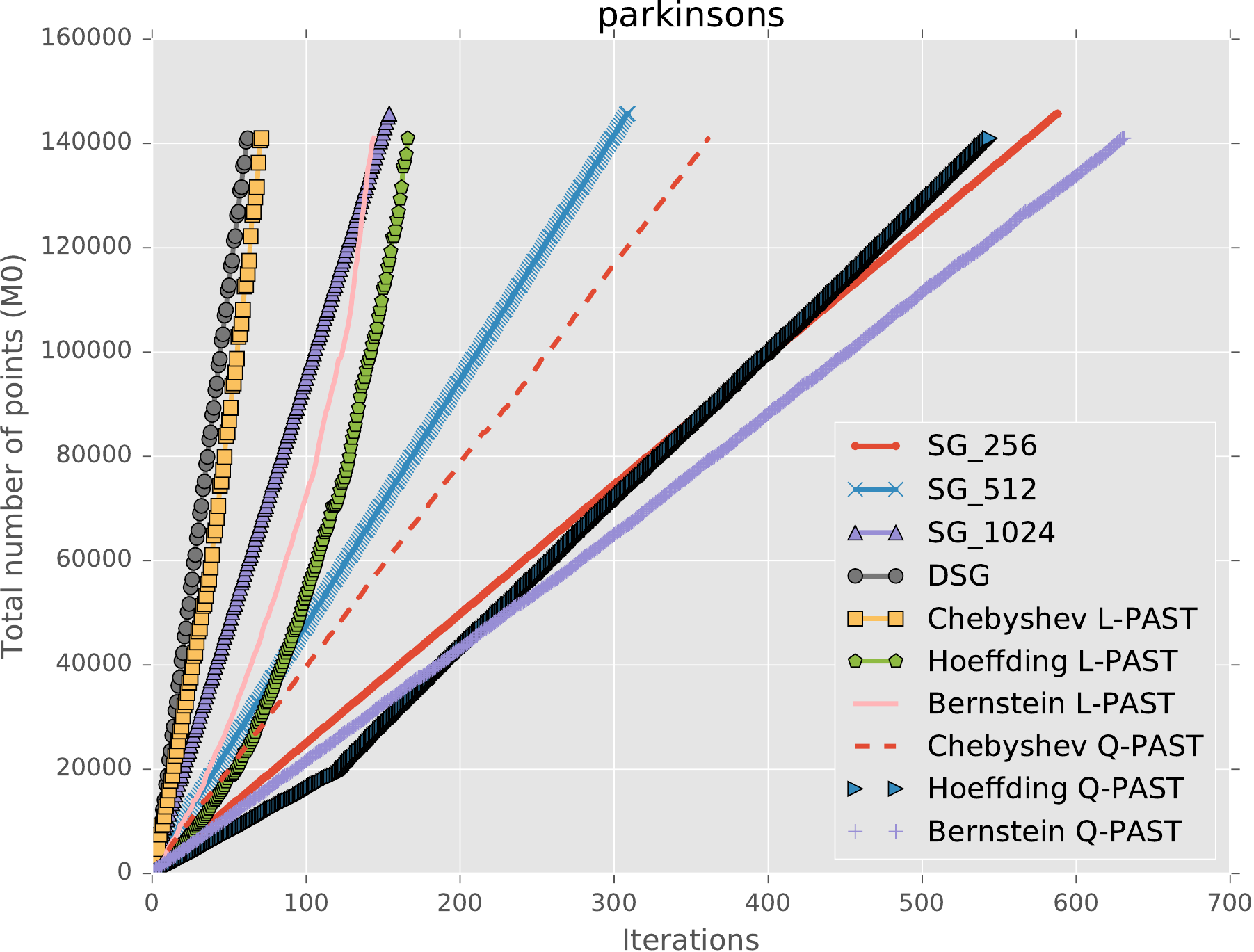}
 \caption{Parkinsons dataset. R2-score and total number of samples. The algorithms have been trained for $30$ epochs.}
 \label{F:regression}
\end{figure}

\subsection{Q-PAST Behavior.}
In this section we evaluate the performance of the quadratic approximation on the regression task.
We assume the problem to be separable in order to consider only the diagonal component of the Hessian.
Figure~\ref{F:regression} shows that Q-PAST outperforms all the other approaches.
It is worth to notice that it is less aggressive in the change of the batch dimension in particular when compared with L-PAST (bottom figure).

Table~\ref{T:levels-r2} shows the R2-score achieved by the algorithm with different confidence levels. It is possible to observe that the influence of $\delta$ on the final performance is very limited. This means that the design of such value is not critical.

\begin{table}[t]
 \centering
 \begin{tabular}{cccccccc}
 \hline
$\delta$&
DSG&
C. L-PAST&
H. L-PAST&
B. L-PAST&
C. Q-PAST&
H. Q-PAST&
B. Q-PAST\\
\hline
$0.1$&
$0.0932$&
$0.0980$&
$0.1350$&
$0.1303$&
$0.1443$&
$0.1451$&
\textbf{$0.1456$}\\
$0.2$&
$0.1083$&
$0.1242$&
$0.1391$&
$0.1344$&
$0.1448$&
\textbf{$0.1460$}&
$0.1458$\\
$0.5$&
$0.1227$&
$0.1364$&
$0.1381$&
$0.1374$&
$0.1462$&
$0.1452$&
\textbf{$0.1466$}\\
 \hline
 \end{tabular}
 \caption{Regression taks. R2-score with different confidence levels. SG-256, SG-512 and SG-1024 obtained
 $0.1449$, $0.1419$ and $0.1330$, respectively. The algorithms have been trained over $30$ epochs.}
 \label{T:levels-r2}
\end{table}

\subsection{Non-Stationary Scenario}
PAST approaches regulate the batch dimension accordingly to the statistical information associated to the estimated gradient.
In particular, when we are far away from the optimal solution we can exploit noisy steps (\ie small batch) to rapidly approach ``good'' solution.
Instead, when we approach the optimal solution we need an accurate estimate of the gradient to closely converge to the optimum.

This property is relevant in realistic applications (\eg online scenario) where the optimal solution may change (even drastically) overtime.
To simulate this scenario we have considered a regression problem ($2$-degree polynomial) where the optimal solution is changed every $35$ iterations (\ie parameter updates).

\begin{figure}[h]
\centering
\includegraphics[width=0.6\columnwidth]{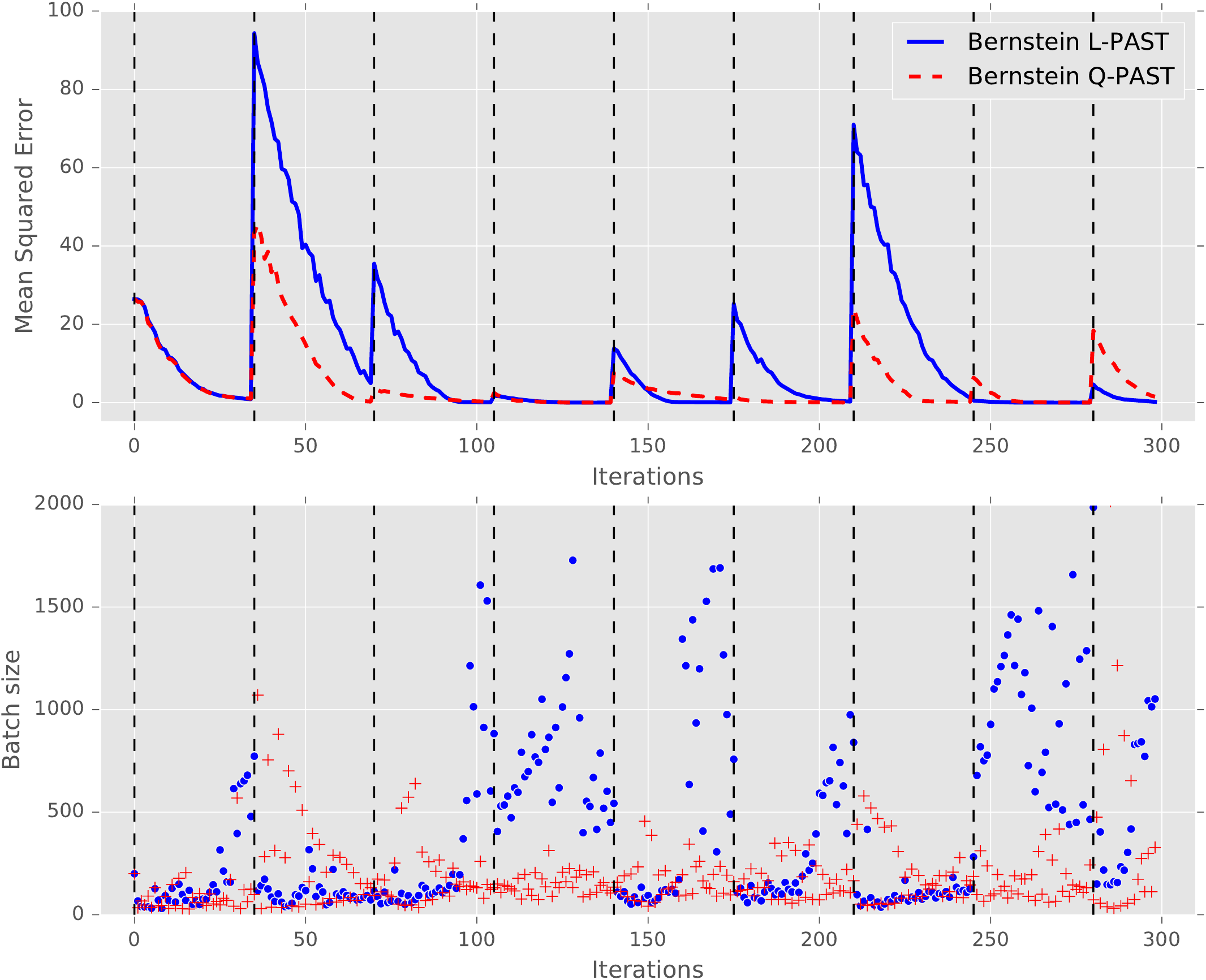} 
\includegraphics[width=.59\columnwidth]{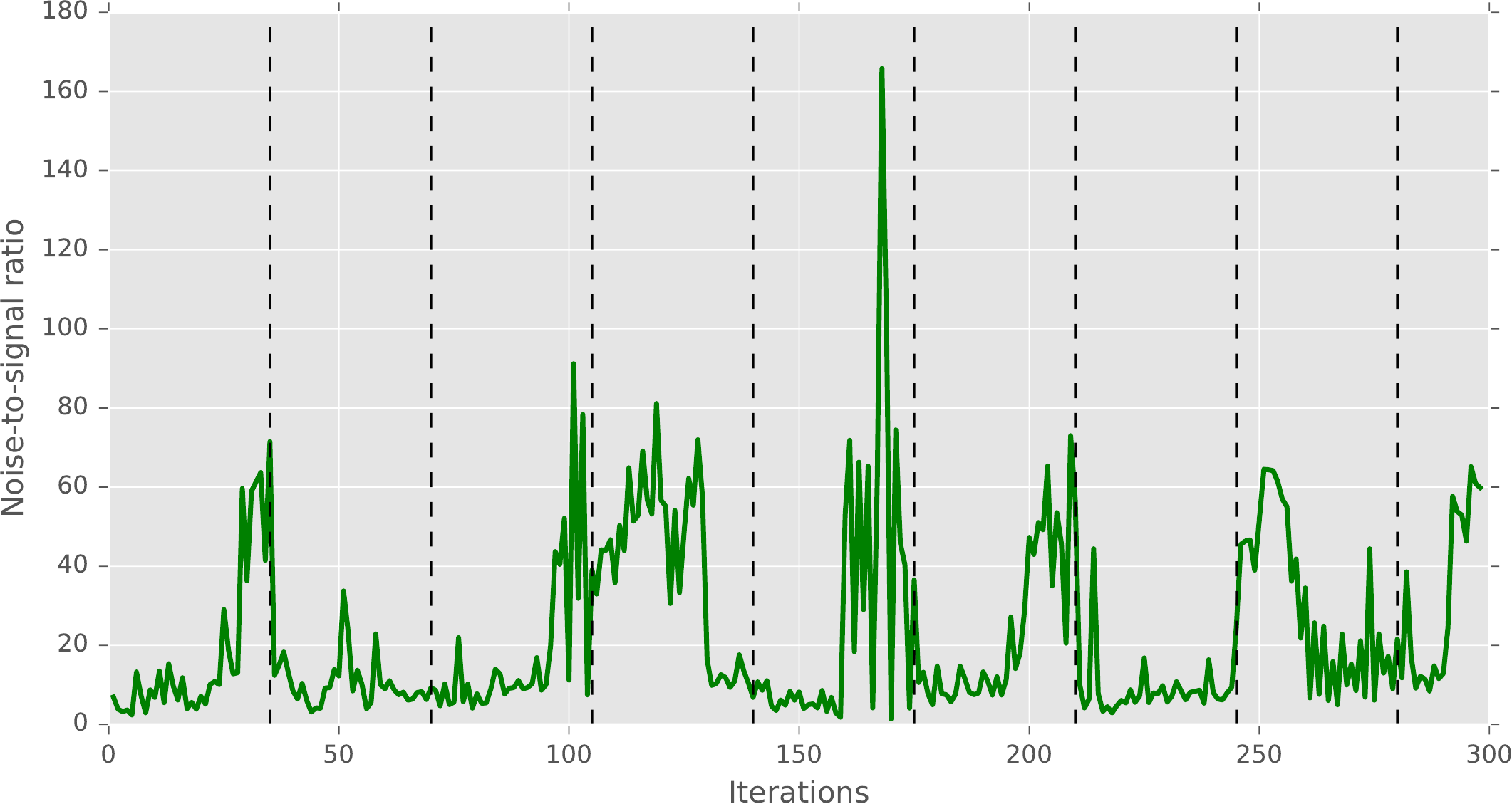} 
\caption{Bernstein L-PAST and Q-PAST in a non-stationary scenario. The model function $f$ is a noisy $2$-degree polynomial where the coefficients changes every $35$ iterations (dashed vertical line).
Above: MSE as a function of iteration, center the corresponding batch size, and, below the noise-to-signal ratio that is connected to the batch size selected by B. L-PAST.}
\label{F:bowl_nostat}
\end{figure}

Figure~\ref{F:bowl_nostat} shows how Bernstein L-PAST handles such scenario.
Initially, small batches are exploited to approach the optimum, then the batch size is increased proportionally to the noise-to-signal ratio.
Intuitively, when the noise-to-signal ratio is high we need to average the gradient over many samples in order to lower the influence of the noise.
A good proxy for the noise-to-signal ratio is provided by the ratio between variance and squared Euclidean norm of the gradient ($\frac{\norm[1]{\text{Var}[\nabla_{\vtheta}\loss^n]}}{\norm[2]{\nabla_{\vtheta}\loss^n}^2}$), see Figure~\ref{F:bowl_nostat}.
When the optimum is changed the algorithm detects a decrease of the noise-to-signal variance (the gradient norm increases \wrt the variance) and adapts the batch size to the new scenario.
This analysis is even more clear when we consider Chebyshev L-PAST since it directly optimize the batch size accordingly to the noise-to-signal ratio.
On contrary, Hoeffding L-PAST, which is less aware of the informed that the other approaches, considers only the gradient magnitude.

In the same figure it is reported the performance of Bernstein Q-PAST. It is possible to observe that Q-PAST selects smaller batches than L-PAST.
This means that it performs noisy steps that leads to less overfitting \wrt to L-PAST.
In fact, when a change in the objective happens, Q-PAST enjoy lower losses than L-PAST.

\section{Conclusions}\label{S:Conclusions}
Pure SG has proved to be effective in several applications, but it is highly time consuming since it exploits one sample for each update.
We have shown that it is possible to exploit automatic techniques that are able to adapt the batch size overtime.
Moreover, these techniques can be used in conjunction to any schema for the update of the parameters.
While L-PAST based on Bernstein's inequality has proved to be effective on the well known MNIST task and Reuters dataset, Q-PAST has proved to be more effective in the regression problem.
However, the computation or estimation of the Hessian may be prohibitive in big data applications such as deep neural networks.

Although the batch size may not play a fundamental role in supervised applications, it is a critical parameters in reinforcement learning specially when the environment is highly stochastic (update the estimate with one sample may be too optimistic).
Future work will apply the proposed techniques to refine policy gradient approaches.

\clearpage

\clearpage

\clearpage
\bibliography{nips16.bib}
\bibliographystyle{plainnat}

\clearpage
\section{Appendix}
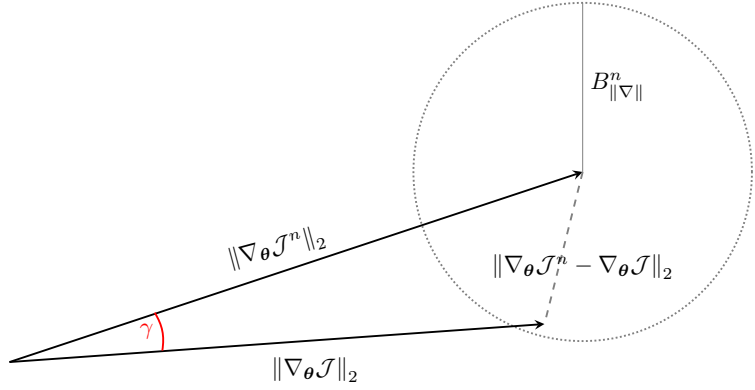
\begin{figure}[t]
\centering
\begin{tikzpicture}[scale=0.9]
\tikzset{>=stealth}
    \begin{axis}[
    hide axis,
    xmax = 2,
    ymax = 1.2,
    height = 9cm,
    width = 14cm,
    thick
    ]
     \addplot [->] plot coordinates {(0,0) (1.5,0.5)};
     \draw[densely dotted, gray] (axis cs:1.5,0.5) circle (2.5cm);
     \draw[gray,thin] (axis cs:1.5,0.5) -- ++(0,2.5cm) node[black,midway,xshift=15pt]{$B_{\norm[]{\nabla}}^{n}$};
     \node[rotate=15] at (axis cs: 0.7,0.3) {$\norm[2]{\nabla_{\vtheta}\loss^n}$};

     \addplot [->] plot coordinates {(0,0) (1.4,0.1)};
     \node at (axis cs: 0.8,-0.02) {$\norm[2]{\nabla_{\vtheta} \loss}$};

     \addplot [-,gray,dashed] plot coordinates {(1.5,0.5) (1.4,0.1)};
     \node at (axis cs: 1.5,0.25) {$\norm[2]{\nabla_{\vtheta}\loss^n -\nabla_{\vtheta}\loss}$};

     \path [-,red] (axis cs:0.38,0.13) edge [bend left=20] node[midway,left] {$\gamma$} (axis cs:0.4,0.03);
    \end{axis}
\end{tikzpicture}
\caption{Graphical representation of the concentration inequality related to the estimated gradient.}
\label{F:vectorangle}
\end{figure}

\subsection{Bounding the Expected Improvement with Global Step Size}
Consider the quadratic expansion of the expected improvement:
\begin{equation}
\label{E:qei}
\simpimp = \transpose{\nabla_{\vtheta}\loss} \Delta\vtheta^n + \frac{1}{2} \transpose{\Delta\vtheta^n} H_{\vtheta}\loss \Delta\vtheta^n,
\end{equation}
where $\Delta\vtheta^n = \lstep \nabla_{\vtheta} \loss^n$ and $\lstep$ is a scalar value.
In this section we will show how to bound the linear and quadratic terms of the expansion.

\subsection{Bounding the Linear Term of $\eimp$}
\label{S:lb_sample_improvement}
In the following we will show how to bound the expected improvement obtained through stochastic gradient update, by means of an upper confidence bounds for the estimated gradient.

Consider the generic representation of the expected and estimated gradient vectors provided in Figure~\ref{F:vectorangle}.
By fixing a confidence value $\delta$ and a concentration inequality, we can write that:
\begin{equation}
\label{E:concentration_bound_squared}
\norm[2]{\nabla_{\vtheta}\loss - \nabla_{\vtheta}\loss^n} < B_{\norm[]{\nabla}}^{n}, \quad \text{\wp } 1-\delta.
\end{equation}
From previous inequality, by noticing that the exact gradient must lie in the ball of radius $B_{\norm[]{\nabla}}$ around the estimated gradient, it is easy to derive the following relationships:
\begin{equation}
\label{E:gnorm_ineq}
\norm[2]{\nabla_{\vtheta}\loss^n} - B_{\norm[]{\nabla}}^{n} < \norm[2]{\nabla_{\vtheta}\loss} < \norm[2]{\nabla_{\vtheta}\loss^n} + B_{\norm[]{\nabla}}^{n}.
\end{equation}

Exploiting a simple trigonometric relationship and inequalities~\eqref{E:concentration_bound_squared}--\eqref{E:gnorm_ineq}, we can write that, \wp $1-\delta$:
\begin{align*}
\left(B_{\norm[]{\nabla}}^n \right)^2
&> \norm[2]{\nabla_{\vtheta} \loss - \nabla_{\vtheta} \loss^n}^2\\
&= \norm[2]{\nabla_{\vtheta} \loss}^2 + \norm[2]{\nabla_{\vtheta} \loss^n}^2 - 2 \norm[2]{\nabla_{\vtheta} \loss} \norm[2]{\nabla_{\vtheta} \loss^n} \cos\gamma\\
&> \left( \norm[2]{\nabla_{\vtheta} \loss^n} - B_{\norm[]{\nabla}}^{n} \right)^2 + \norm[2]{\nabla_{\vtheta} \loss^n}^2 - 2 \norm[2]{\nabla_{\vtheta} \loss} \norm[2]{\nabla_{\vtheta} \loss^n} \cos\gamma.
\end{align*}
Then,
\begin{equation*}
\cos\gamma > \frac{\norm[2]{\nabla_{\vtheta} \loss^n} - B_{\norm[]{\nabla}}^{n}}
{\norm[2]{\nabla_{\vtheta} \loss}},\quad \text{\wp } 1-\delta.
\end{equation*}

As a consequence, the linear term of~\eqref{E:qei} can  be lower bounded as follows (\wp $1-\delta$)
\begin{align*}
\transpose{\nabla_{\vtheta} \loss} \nabla_{\vtheta}\loss^n = \norm[2]{\nabla_{\vtheta} \loss} \norm[2]{\nabla_{\vtheta} \loss^n} \cos\gamma
> \left( \norm[2]{\nabla_{\vtheta} \loss^n} - B_{\norm[]{\nabla}}^{n} \right) \norm[2]{\nabla_{\vtheta} \loss^n}
\end{align*}

Finally, \wp $1-\delta$
\begin{equation*}
 \simpimp_L = \transpose{\nabla_{\vtheta}\loss} \Delta\vtheta^n >
 \plb_{L}^n(\lstep) = \lstep \left( \norm[2]{\nabla_{\vtheta} \loss^n} - B_{\norm[]{\nabla}}^{n} \right) \norm[2]{\nabla_{\vtheta} \loss^n}.
\end{equation*}

\subsection{Bounding the Quadratic Term of $\imp^S$}
Let us consider the quadratic term in~\eqref{E:qei}.
Then:
\begin{align*}
\transpose{\nabla_{\vtheta}\loss^S} H_{\vtheta}\loss \nabla_{\vtheta}\loss^S
&= \sum_{i,j} H_{\vtheta}^{(ij)}\loss \nabla_{\vtheta}\loss^S_i \nabla_{\vtheta}\loss^S_j
\end{align*}
Previous equation requires the knowledge of the exact Hessian in the current parametrization.
However, we can only estimate such quantity from observations.
In order to derive a lower bound to such quadratic term, we assume that the following inequalities hold \wp $1-\delta$:
\begin{equation}
 \left| H_{\vtheta}^{(ij)} \loss - H_{\vtheta}^{(ij)} \loss^S \right| < B_H^{(ij)} \quad \forall i,j.
\end{equation}
In order to properly handle the approximation error and obtain a lower bound we need just to subtract the bound:
\begin{align*}
\transpose{\nabla_{\vtheta}\loss^S} H_{\vtheta}\loss \nabla_{\vtheta}\loss^S
&> \sum_{i,j} \left( H_{\vtheta}^{(ij)}\loss - B_H^{(ij)} \right) \nabla_{\vtheta}\loss^S_i \nabla_{\vtheta}\loss^S_j, \quad{} \text{\wp \;} 1-\delta
\end{align*}
then, \wp $1-\delta$:
\begin{align*}
\simpimp_Q 
&= \transpose{\nabla_{\vtheta}\loss} \Delta\vtheta^S + \frac{1}{2} \transpose{\Delta\vtheta^S} H_{\vtheta}\loss \Delta\vtheta^S\\
&> \lstep \left( \norm[2]{\nabla_{\vtheta} \loss^S} - B_{\norm[]{\nabla}}^{n} \right) \norm[2]{\nabla_{\vtheta} \loss^S}
+ \frac{1}{2} \lstep^2 \sum_{i,j} \left( H_{\vtheta}^{(ij)}\loss - B_H^{(ij)} \right) \nabla_{\vtheta}\loss^S_i \nabla_{\vtheta}\loss^S_j.
\end{align*}

\section{Probabilistic Adaptive Sample Technique}
In this section we derive the Probabilistic Adaptive Sample Technique (PAST) used to select both the batch size $n$ and the step size $\lstep$.
First off all we want to review the concentration inequalities. 

%

We start the Hoeffding's inequality for uncentered random vectors.
The following lemma is adapted from~\citep[Theorem 1.3]{tropp2012user}.

\begin{lemma}[Vector Hoeffding]
Consider a finite sequence $\{\mathbf{x}_k\}$ of independent, random vectors with dimension $d$ and population variance $\boldsymbol{\mu} = \mathbb{E}[\mathbf{x}]$, 
such that, for any $k$, $\norm[2]{\mathbf{x}_k} \leq L$.
Introduce the mean
$\mathbf{Z} = \frac{1}{n} \sum_k \mathbf{x}_k,$ 
then, for all $t\geq 0$,
$$P\left( \norm[2]{\mathbf{Z} - \boldsymbol{\mu}} \geq t \right)
\leq (d + 1)\cdot e^{-\frac{nt^2}{8 L^2}}.
$$
\end{lemma}

Note that Hoeffding's inequality only requires the knowledge of the support of the distribution (here $L$).
On contrary, Chebyshev's and Bernstein's inequalities require in addition the knowledge of the distribution variance.

We start considering the matrix version of the Chebyshev's equality provided in~\citep{ferentios1982chebyshev}.
Here we report the Chebyshev's inequality for the sample mean.

\begin{lemma}[Vector Chebyshev]
Consider a finite sequence $\{\mathbf{x}_k\}$ of independent random vectors with common dimension $d$. 
Let $\boldsymbol{\mu}$ be the population mean and $\nu(\boldsymbol{x}) = \transpose{\left[\sigma_1^2,\ldots,\sigma_{\pdim}^2\right]}$ the vector variance statistic storing the population variance of each component.
Introduce the mean
$\mathbf{Z} = \frac{1}{n} \sum_k \mathbf{x}_k,$
then, for all $t \geq 0$,
$$
P\left(
\norm[2]{\boldsymbol{Z} - \boldsymbol{\mu}} \geq 
\frac{
t \norm[2]{\sqrt{\nu(\boldsymbol{x})}}
}{\sqrt{n}}
\right)
\leq \frac{1}{t^2}.
$$
\end{lemma}

\begin{lemma}[Vector Bernstein, adapted from (\citealt{tropp2015introduction}, Corollary 6.1.2)]
Consider a finite sequence $\{\mathbf{x}_k\}$ of independent random vectors with common dimension $d$. 
Let $\boldsymbol{\mu}$ be the population mean and $\nu(\boldsymbol{x}) = \transpose{\left[\sigma_1^2,\ldots,\sigma_{\pdim}^2\right]}$ the vector variance statistic storing the population variance of each component.
Assume that each vector is such that $\norm[2]{\mathbf{x}_k} \leq L$, \ie each vector has uniformly bounded deviation from the mean:
$$\norm[2]{\mathbf{x}_k - \boldsymbol{\mu}}\leq L, \quad \forall k.$$
Introduce the mean
$\mathbf{Z} = \frac{1}{n} \sum_k \mathbf{x}_k,$
then, for all $t \geq 0$,
$$P\left(
\norm[2]{\mathbf{Z} - \boldsymbol{\mu}} \geq t
\right)
\leq (d+1) \; \exp \left(\frac{-nt^2/2}{
\norm[2]{
\nu(\mathbf{Z})
}
+Lt/3}\right).
$$
\end{lemma}

\subsection{Linear-PAST}
Given the lower bound $\plb_{L}^n(\lstep)$ derived in Section~\ref{S:lb_sample_improvement} we need to solve the cost sensitive problem in~\eqref{E:costproblem}, reported here for sake of clarity
\begin{equation*}
 n' 
 = \argmax_{\bar{n} \in \mathbb{N}^+} \frac{\lstep \left( \norm[2]{\nabla_{\vtheta} \loss^n} - B_{\norm[]{\nabla}}^{\bar{n}} \right) \norm[2]{\nabla_{\vtheta} \loss^n}}{\bar{n}},
\end{equation*}
where $\nabla_{\vtheta} \loss^n$ is the realization of the random variable computed on a subset of $S$ of dimension $n$ (it is independent from the value $\bar{n}$).
It is easy to observe that the problem can be further reduced
\begin{equation*}
 n' 
 = \argmax_{\bar{n} \in \mathbb{N}^+} \frac{\norm[2]{\nabla_{\vtheta} \loss^n} - B_{\norm[]{\nabla}}^{\bar{n}}}{\bar{n}}.
\end{equation*}
The bound $B_{\norm[]{\nabla}}^{n}$ as to be replaced using one of the concentration inequalities presented above.
\begin{itemize}
 \item Hoeffding's inequality: \wp $1-\delta$
 $$\norm[2]{\nabla_{\vtheta} \loss^n - \nabla_{\vtheta} \loss} \leq L\sqrt{ \frac{8}{n} \ln\left(\frac{\pdim +1}{\delta}\right) },$$
 then
 $$n \geq \frac{18 L^2}{\norm[2]{\nabla_{\vtheta}\loss^S}^2} \ln\left(
\frac{\pdim+1}{\delta}
\right).$$
 \item Chebyshev's inequality: \wp $1-\delta$
 $$\norm[2]{\nabla_{\vtheta} \loss^n - \nabla_{\vtheta} \loss} \leq \sqrt{\frac{\norm[2]{\sqrt{\nu(\nabla_{\vtheta}\loss)}}^2}{\delta n}} = \sqrt{\frac{\norm[1]{\nu(\nabla_{\vtheta}\loss)}}{\delta n}},$$
 then
 $$n \geq \frac{9\norm[1]{\nu(\nabla_{\vtheta}\loss)}}{4\delta \norm[2]{\nabla_{\vtheta} \loss^n}^2}.$$
 \item Bernstein's inequality: \wp $1-\delta$
 $$\norm[2]{\nabla_{\vtheta} \loss^n - \nabla_{\vtheta} \loss} \leq \sqrt{
 \frac{2 \norm[2]{\nu(\nabla_{\vtheta}\loss)}}{n}\ln\left(\frac{d+1}{\delta}\right)
 }
 + 
 \frac{2 L}{3n}\ln\left(\frac{d+1}{\delta}\right),
 $$
 then
 $$n \geq \frac{9b+16a\norm[2]{\nabla_{\vtheta} \loss^n}+3\sqrt{9b^2+32ab \norm[2]{\nabla_{\vtheta} \loss^n}}}{8\norm[2]{\nabla_{\vtheta} \loss^n}^2},$$
 where
 $$a := \frac{2}{3} L \ln\left(\frac{d+1}{\delta}\right), \qquad b := 2 \norm[2]{\nu(\nabla_{\vtheta}\loss)}\ln\left(\frac{d+1}{\delta}\right).$$
\end{itemize}

\end{document}